\documentclass[11pt,a4paper]{article}
\usepackage{naaclhlt2019}
\usepackage{times}
\usepackage{latexsym}
\usepackage[htt]{hyphenat}
\usepackage{relsize}

\usepackage{url}
\usepackage{todonotes}
\usepackage{float}

\usepackage{graphicx}
\graphicspath{figures}
\usepackage{caption} 
\usepackage[labelformat=simple]{subcaption}

\aclfinalcopy % Uncomment this line for the final submission
\title{Chat-crowd: A Dialog-based Platform for Visual Layout Composition}

\author{Paola Cascante-Bonilla$^1$ \quad Xuwang Yin$^1$ \quad Vicente Ordonez$^1$  \quad Song Feng$^2$ \\
  $^1$University of Virginia, $^2$IBM Thomas J. Watson Research Center.\\
  {\tt [pc9za, xy4cm, vicente]@virginia.edu, sfeng@us.ibm.com }
  }

\date{}

\begin{document}
\maketitle
\begin{abstract}
  In this paper we introduce Chat-crowd, an interactive environment for visual layout composition via conversational interactions. Chat-crowd supports multiple agents with two conversational roles: agents who play the role of a \emph{designer} are in charge of placing objects in an editable canvas according to instructions or commands issued by agents with a \emph{director} role. The system can be integrated with crowdsourcing platforms for both synchronous and asynchronous data collection and is equipped with comprehensive quality controls on the performance of both types of agents. We expect that this system will be useful to build multimodal goal-oriented dialog tasks that require spatial and geometric reasoning. 
\end{abstract}

\section{Introduction}
\label{sec:introduction}
There has been growing interest in building visually grounded dialog systems~\cite{ren2015exploring,bisk2016natural,das2017visual,chen2018touchdown,el2018keep,shridhar2018interactive}. Building interactive agents that can complete goal-oriented tasks in a situated environment using natural language is a challenging problem that requires both robust natural language understanding (NLU) and natural language generation (NLG). 
%For instance, a task of human-robot interactions requires NLU to convert language into actionable procedures that the robot can execute. As language is inherently ambiguous, the robotic agent should likewise be able to interact back using NLG to request disambiguating or clarifying instructions. 
Datasets for visually grounded dialog tasks have started to emerge but more general and effective tools for data collection are still missing.  %Various aspects of this problem are partially addressed in recently proposed tasks such as human robot communication \cite{bisk2016natural},
% image captioning~\cite{chen2015microsoft}
%visual question answering~\cite{ren2015exploring}, and more recently, visually grounded dialogs~\cite{das2017visual}. 
% While these previous efforts leverage real-world images, this also leads to the language not being as complex, and simple baselines have shown remarkable performance in some of these tasks (e.g.~\citet{devlin2015exploring,jabri2016revisiting}).

\begin{figure}[t]
\centering
\includegraphics[width=0.85\linewidth]{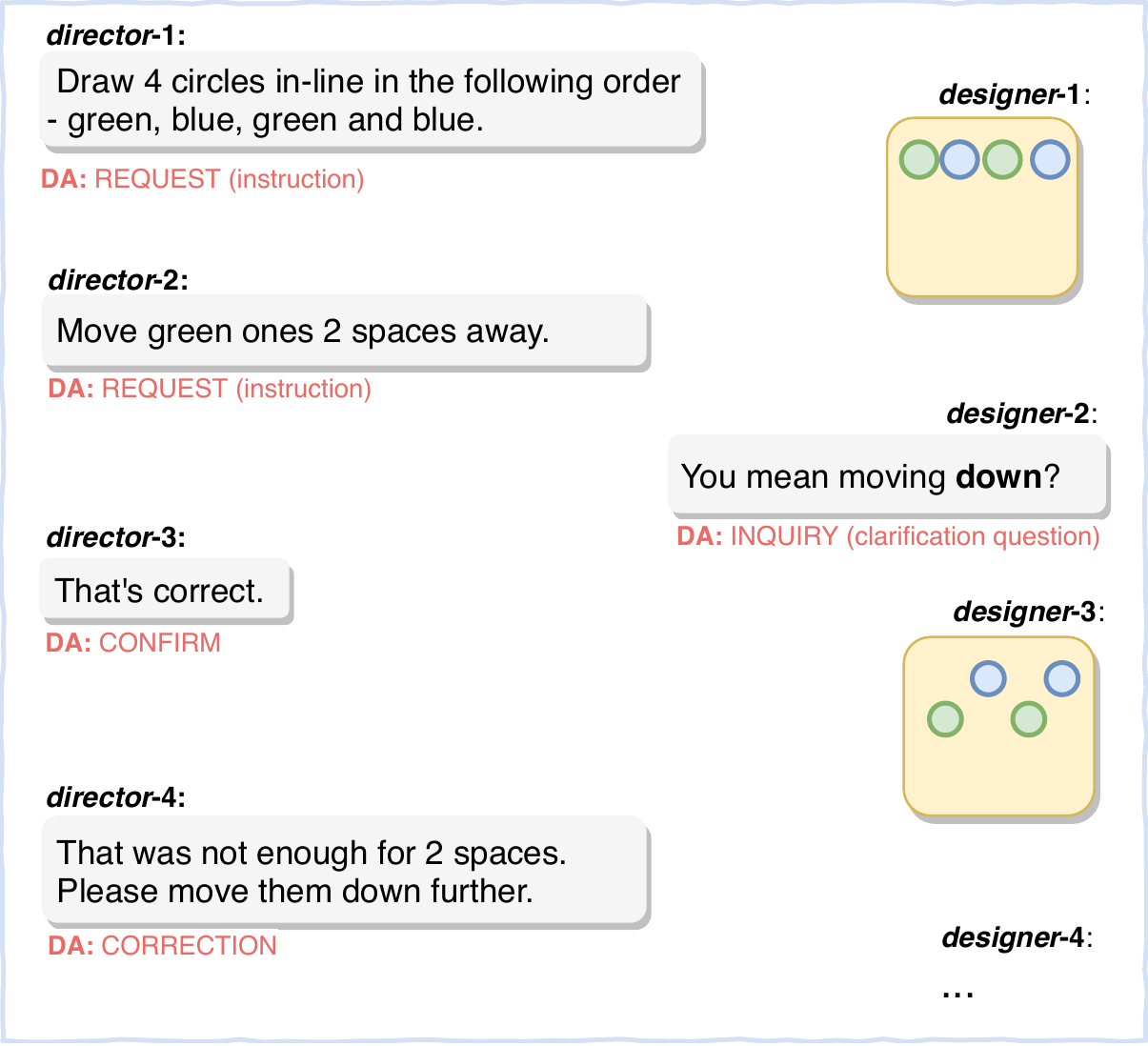}
% \vspace{-0.05in}
\caption{An illustration of the interactions and dialog acts (DA) between a \emph{director} and a \emph{designer} for one of our sample tasks. In asynchronous mode, the role of an agent can be taken by a different user in each round.
%A dialog between a \emph{director} and a \emph{designer} agent for one of our sample tasks. In asynchronous mode, the role of an agent can be taken by a different user in each round. 
%The dialog is ended when the original layout presented to the \emph{director} is matched. Note: we include screenshots of the actual user interface used for the \emph{director} and \emph{designer} in the supplementary material. An illustration of the dialog interactions and  acts (DA) between a \emph{director} and a \emph{designer} agent.
}
\vspace{-0.2in}
\label{fig:main-figure}
\end{figure}

\renewcommand*{\UrlFont}{\ttfamily\smaller}

We introduce an interactive data collection and annotation tool\footnote{ \url{chatcrowd.github.io}} for the collaborative tasks of visual layout composition through natural language dialogs (see Figure~\ref{fig:main-figure}). In this work, we refer to layouts to the spatial distribution of objects in a 2D canvas as well as their attributes such as name, shape, or color. More specifically, Chat-crowd is designed to support a basic model task consisting of dialogs between a \emph{director} agent that is provided with a visual layout as a reference, and a \emph{designer} agent that is provided with a canvas where one can add, remove, resize, or relocate visual elements. The two agents can communicate using natural language within the context of various dialog acts (``DA'') as illustrated in Figure \ref{fig:main-figure}. The \emph{director} provides instructions to modify elements in the canvas, and the \emph{designer} optionally writes clarifying questions or modifies the canvas. While the \emph{director} can see the progress of the \emph{designer} as the dialog proceeds, the \emph{designer} only sees the instructions from the \emph{director}. The dialog ends when the visual layout composed by the \emph{designer} matches the given visual layout to the \emph{director}. 

One key feature of our system is that it allows asynchronous conversations, i.e. the \emph{director} and \emph{designer} do not need to be online at the same time, or be persistent throughout the task. This means that different users can pick up the task where it was left off in the previous interaction, thus simplifying the overall collection process. Furthermore, it enables a process for data validation and job distribution at a finer level. Additionally, our system optionally employs a bot agent to inject synthetic utterances that trigger diversified or less represented dialog activities. Such injections can also be used for evaluating the responses from human agents since the optimal responses to the synthetic utterances are already given. We show that under this asynchronous mode, crowdsourced contributors are still able to converse based on the dialog history and complete the tasks. 

% Our system is designed to enable a large coverage and diversity of geometrical, spatial, and semantic relations between objects in both the visual and language modalities. In particular, our system supports the generation of diverse dialog acts (e.g., \textsc{request}, \textsc{confirm}, \textsc{inquiry}, and \textsc{correction}), which are typically encountered in the long tail in dialog datasets \cite{stolcke2000dialogue} due to the uncontrolled conditions in a crowdsourcing platform. 
% Our annotation tool supports multimodal conversational data collection for two sample visual layouts: the layouts corresponding to abstract scenes with simple geometrical primitives, and the layouts corresponding to real-world scenes consisting of the locations for physical objects.

% Since our inputs are layouts of objects, the dialogs are not grounded on pixel images but rather abstract pictorial representations. 
To validate our system, we first apply the system on a task with more controllable visual layouts consisting of visual primitives (e.g., circles, rectangles, triangles); we also test our system on grounding for visual layouts corresponding to objects in real-images from an object recognition dataset \cite{lin2014microsoft}. Separating the pattern recognition task from the visual understanding task through visual layouts allows us to explore richer language for spatial reasoning yet still connects to real-world images. 
% Additionally, exploring these two settings allows us to still connect our work on applications to real-world images. 

Our contributions are the following: (1) a new multi-modal dialog simulation system with a focus on spatial reasoning; (2) an asynchronous dialog collection platform that can trigger more diverse dialog activities and evaluate the performance of crowdsourced contributors in ongoing tasks; (3) an analysis of the difficulty and the type of language used by people to accomplish the proposed collaborative task of re-constructing visual layouts from asynchronous dialogs.

\section{Visual Layout Dialog Collection}
\label{sec:datacollection}
%Our first thrust in this work is to introduce Chat-crowd, a platform for collecting dialog data grounded on visual layouts for goal oriented tasks. 
% We introduce two model tasks to 
This work aims to demonstrate the usage of Chat-crowd for obtaining dialog data for geometric and spatial reasoning, ranging from abstract to more complex scenes. To this end, we explore two types of visual layouts: layouts in a shape-world with automatically generated simple 2D shape primitives, and layouts of objects from real images.% from the COCO dataset~\cite{lin2014microsoft}.

\begin{figure*}[t]
\centering
\begin{minipage}{0.5\textwidth}
 \centering
 \includegraphics[width=.90\linewidth]{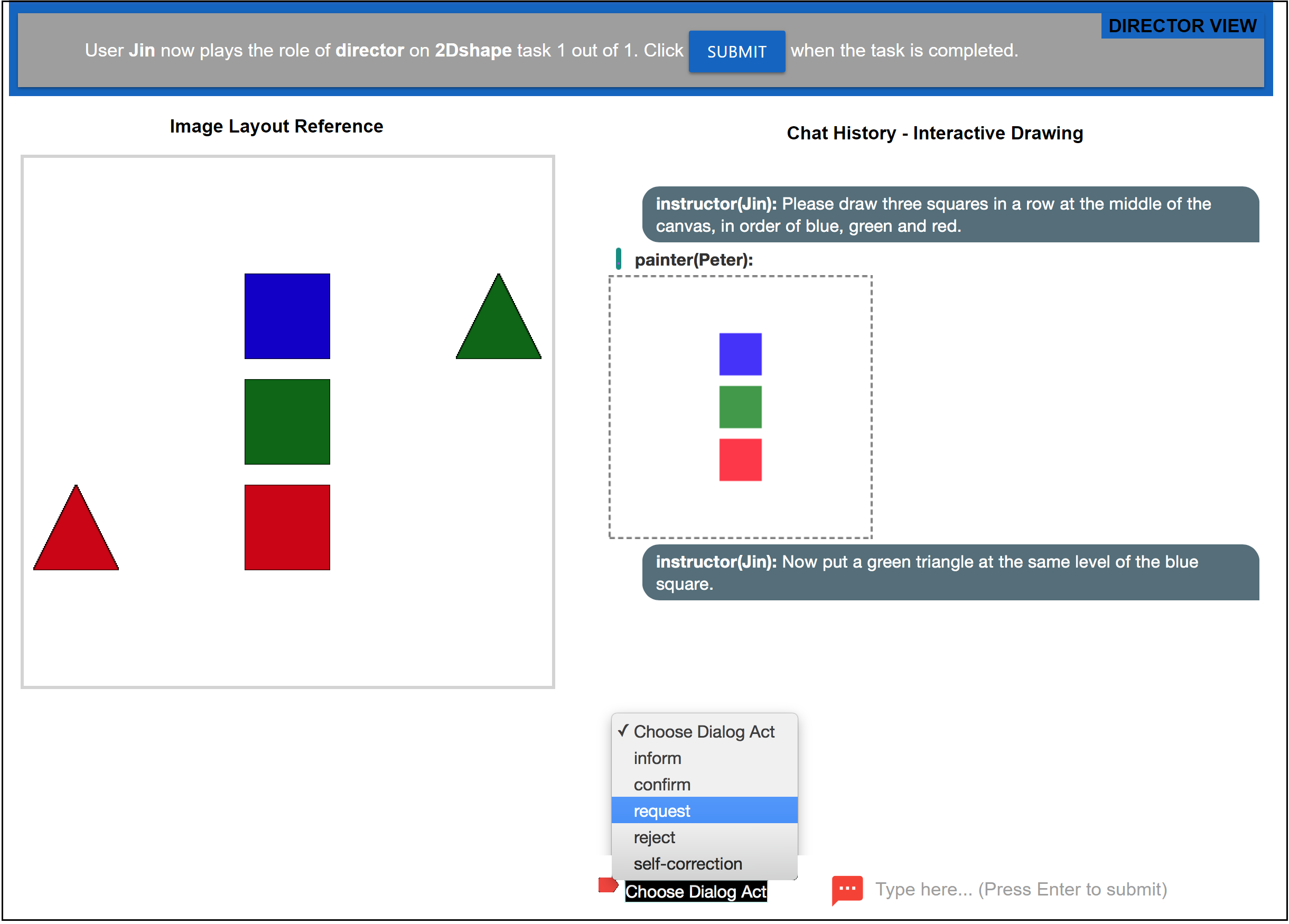}
 \label{fig:director_UI}
\end{minipage}%
\begin{minipage}{0.5\textwidth}
 \centering
 \includegraphics[width=.90\linewidth]{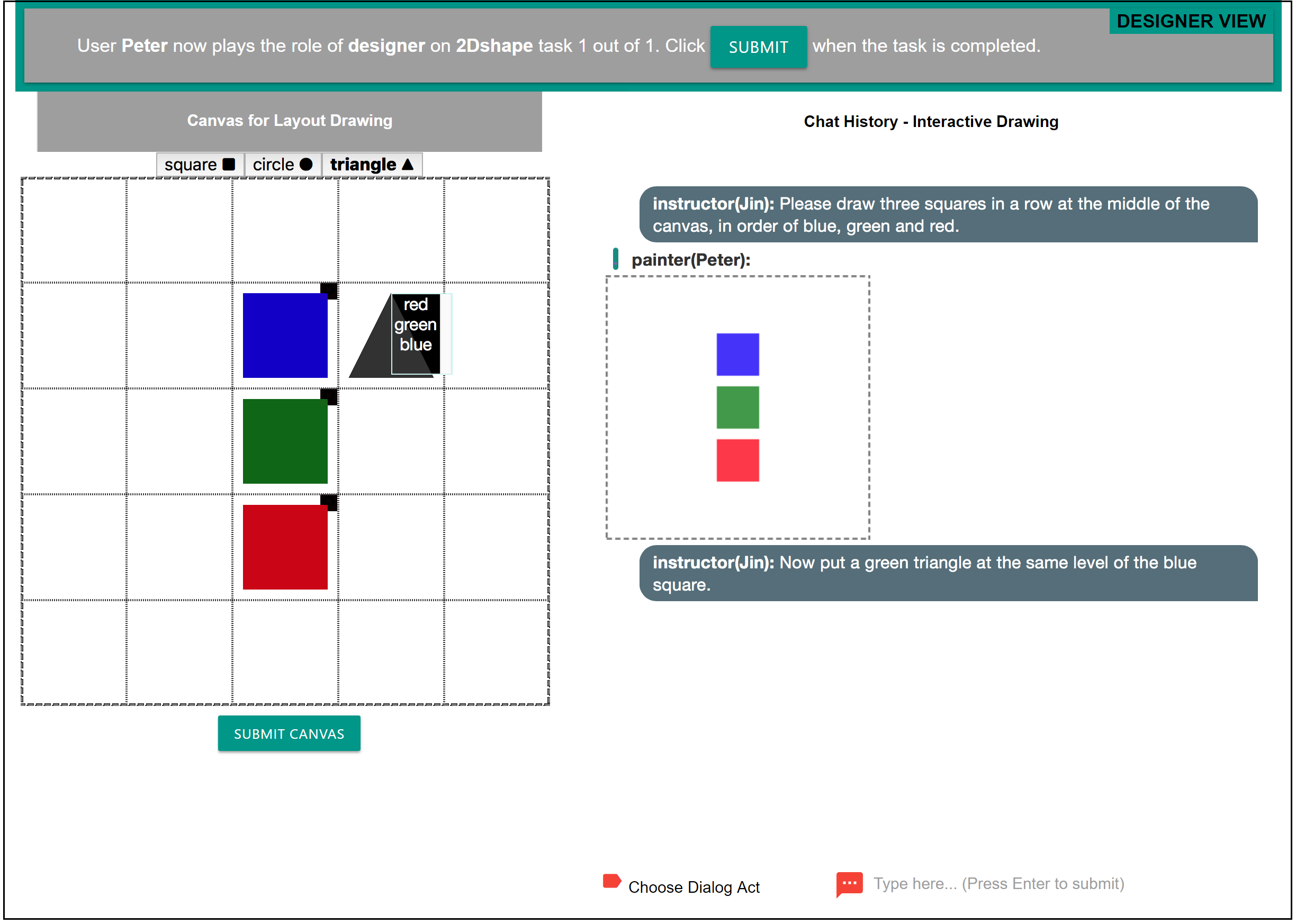}
  \label{fig:designer_UI}
\end{minipage}%
\caption{UI for \emph{director} (left) and \emph{designer} (right) agents for 2D-shape layout task.}
% \vspace{-0.2in}
\label{fig:UI}
\end{figure*}

\vspace{-0.06in}
\paragraph{2D-shape Layouts} We propose a synthetic layout world where objects of different shapes (circle, square, triangle) and colors (blue, red, green) are pinned to a set of $5\times5$ grid locations on a canvas. This setup allows us to focus on the language, and the accurate reconstruction of the visual layouts by discarding the additional complexities of real-world scenes. We generate two types of 2D-shape layouts: (1) \texttt{2d-shape-random}: consisting of scenes with 4 to 6 objects with shapes, color, and locations selected randomly, and \texttt{2d-shape-pattern}: consisting of objects generated by following a set of customizable production rules that encourage adjacent objects.
% : We placed randomly 3 adjacent objects (same row, column or diagonal) with either the same shape or color, and added 2 or 3 more random objects independently generated so that we have a total of 4 to 6 objects as in our previous set of layouts. 
% Note that in these two settings locations are sampled without replacement such that there are no overlapping shapes. The idea of this second type of layout is to analyze whether the patterns in the layout make the task easier. Figure \ref{fig:director_UI} and \ref{fig:designer_UI} 
Figure \ref{fig:UI} presents our user interface for the data collection tasks of 2D-shape layouts. For the real-image layouts, the interface includes additional features for resizing, moving, and naming objects.
% Such layout 2D-shape-random: randomly place 4-6 objects with random shapes and colors on a 5x5 grid. 2D-shape-pattern: a combination of random objects and a random pattern on a 5x5 grid, the total number of objects is 4-6. A pattern is 3 adjacent objects with same shapes or same color, could be aligned in a row, in a column, or in diagonal, could be placed anywhere on the grid, but objects cannot be in different shapes and different color at the same time. 

\vspace{-0.06in}
\paragraph{COCO Layouts} We use as reference and test bed of the object layouts of real-world images from the COCO dataset \cite{lin2014microsoft}. 
% The layout of an image includes 80 classes of objects, and their bounding box locations. 
The layout of an image includes objects and their locations. All objects are represented by a set of rectangles (proportional to the size of the corresponding object) and the object class (e.g., \texttt{people}, \texttt{dog} and \texttt{surfboard}). We also experiment with two types of scenarios: (1) \texttt{COCO-simple}: corresponding to images with simple layouts with 3 to 4 object instances belonging to 3 distinct classes, and (2) \texttt{COCO-complex}: corresponding to images consisting of layouts with 6 to 8 object instances belonging to 6 distinct object classes.

\subsection{Crowdsourcing Task Design} 
% The layouts and empty (or latest) canvases are provided to the crowdsourced contributors who then have to match this vision by communicating with co-workers. 
In our task, crowd agents interact under two roles: \emph{director} and \emph{designer}. 
In the \emph{director} mode, agents direct the drawing in the following ways: (1) providing instructions for how layouts should be modified; (2) giving suggestions for correcting or improving the current layout; (3) answering questions from the \emph{designer} agents. In the \emph{designer} mode, the agents either follow the instructions to draw on the canvas by specifying the attributes and locations for 2D-shapes/COCO, or ask clarifying questions if needed. 

\paragraph{Data Collection} One challenge for such multi-model dialog collection via crowdsourcing is that it could be very complicated and expensive to pair two qualified contributors to converse in real time~\cite{lasecki2013interactive}. Our system is designed to support both synchronous as well as asynchronous interactions. In the asynchronous mode, the agents are asked to review and understand a chat history before taking an action. Thus, we design quizzes to assist agents in learning how to examine the chat history to determine what actions are helpful for reconstructing the layouts. 

\paragraph{Quality Control} The most common quality assurance provided by crowdsourcing platforms is to evaluate the performance with gold standard data, which is not applicable in our case. We propose to verify and ensure the success of the task using the following criteria: a task is considered successful if a layout is reconstructed with high similarity with respect to the reference layout. For 2D shape layouts, a layout similarity is tested by computing an exact match; For COCO layouts the matching is confirmed by the \emph{director} agents who determines when the task should end.
    %\item Instruction evaluation: we check the repetitiveness and informativeness in the content via NLP approaches \todo{how? any reference?} In addition, the instructions from the directors are also implicitly evaluated by the designers as they can report the instruction is under-specific or irrelevant etc..
%\end{itemize}

\paragraph{Output Data} The output includes free-form textual utterances from \emph{director} and \emph{designer} agents annotated with dialog acts, a sequence of drawings on the canvas and the final layouts of images. In our task setting, the dialog data can be potentially divided into sub-dialogs or atomic dialog interactions accordingly.

\begin{figure}[t]
\centering
\includegraphics[width=0.85\linewidth]{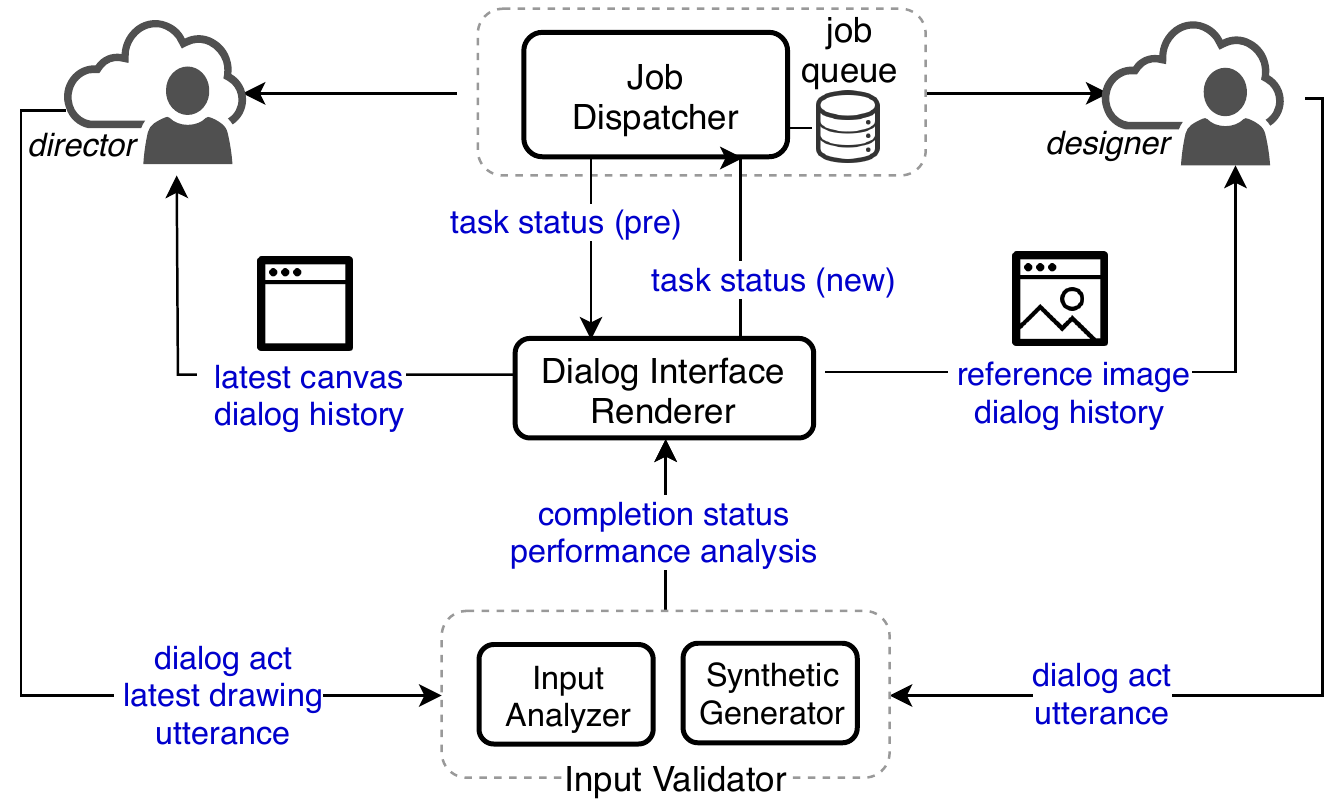}
\caption{Overview of the Chat-crowd System}
\label{fig:archi}
\vspace{-0.2in}
\end{figure}

\section{Experiment Results and Analysis}
\label{sec:experiments}

% In this section we describe additional details of our dialog collection platform.
\subsection{System Overview}
Figure \ref{fig:archi} presents an overview of our system. In synchronous mode, it allows the agents to converse in real time to perform a given task. In asynchronous mode, the automated Job Dispatcher determines a job of a role for next turn to interface with the crowdsourcing platforms. The latest canvas and dialog history of a given job are dynamically generated by Dialog Interface Renderer. Once an input (e.g, dialog act, utterance, or latest canvas) is submitted, Input Validator first examine the content via its sub-module Input Analyzer. It identifies the modification to the previous canvas; object features and locations in the utterance\footnote{We employ NLP tools by  \url{spacy.io}} and the dialog acts etc.. Additionally, Synthetic Generator is applied to inject certain responses: (1) to intrigue more diverse dialog activities such as designer asking clarification questions; (2) to inspect the performance of the contributors, for instance, when a \emph{designer} submits a canvas given a non-viable instruction by Synthetic Generator.

\subsection{Experimental Settings}
We post the jobs on the FigureEight crowdsourcing platform.
% \footnote{Formerly known as CrowdFlower.} 
% Contributor channel is set as Level 3, which means high quality contributors, and English is a requirement for both agents, contributors are further restricted to be in the US, UK, and Canada. Each job consists of 5 layout drawing tasks, and is priced at \$1.20. 
~We collected dialogs for 100 \texttt{2d-shape-random} layouts, 100 \texttt{2d-shape-pattern} layouts, and additionally run a pilot study on 10 \texttt{COCO-simple} layouts and 10 \texttt{COCO-complex} layouts, leading to $2,520$ individual user interactions for 2d-shapes and $595$ for real scene COCO layouts.

\subsection{Quantitative and Qualitative Analysis}

%Analysis on the conversational behavior: deliberative planning, reactive execution? efficiency? language/dialog flow diversity? 
\paragraph{Director Word Usage Analysis} We first analyzed the types of words that people in the \emph{director} agent role used to provide instructions to the \emph{designers}. We found that people mention location, color, and shape words in over $90$\% of the total of instructions and often all three with a slight preference for mentioning shape over color information. For the \texttt{2d-shape-pattern} task there was slightly lower preference to mention shape and color, than in \texttt{2d-shape-random}. This is because when each object is placed randomly, then people have to more often refer to each object individually on each round.

\paragraph{Designer Reactions} Analyzing the interactions by \emph{designer} agents we found that about $60$\% of the times they modify the canvas without necessarily asking clarification questions. Here is a set of example responses: ``I did not understand instructions from instructor ...'', ``please give instruction'', 
examples of questions are: ``where to put circle?'', ``do the boxes mean squares'', ``where exactly, in the middle, left or right?'', ``It is done?''.
%will add statistics

\paragraph{Task Duration} We additionally analyze the difficulty of each task by the number of rounds that it takes to complete a layout, and the average length (in words) for the instructions issued by the \emph{director}. Table~\ref{table:duration_statistics} shows these statistics for our four type of layouts. We found, unsurprisingly, that for \texttt{2d-shape-pattern} layouts, the average number of rounds is significantly lower than for \texttt{2d-shape-random}, indicating that the pattern in the distribution of objects in the canvas is indeed being exploited by the agents. Additionally, we can gauge the difference in difficulty between \texttt{COCO-complex} and \texttt{COCO-simple}, where the number of rounds is more than double even when the average instruction is only two words larger.

\begin{table}[h]
\setlength\tabcolsep{2pt} % default value: 6pt
  \centering
  \begin{tabular}{lcc}
{\sc Layout-type}                                  & \#{\sc Rounds} & {\sc Avg-Words}\\ \hline
\texttt{2d-shape-random} & 7.23     & 18.0 $\pm$ 14.3    \\ 
\texttt{2d-shape-pattern}    & 5.37         & 19.4  $\pm$ 15.8    \\
\texttt{COCO-simple} & 7.6     & 18.7 $\pm$ 13.9 \\ 
\texttt{COCO-complex}  & 22.1    & 20.9 $\pm$ 28.2 \\
  \end{tabular}

  \caption{Statistics for task duration for each type of layout based on the number of rounds needed and average number of words in the instructions.}
  \label{table:duration_statistics}
  \vspace{-0.15in}
\end{table}

\paragraph{Instruction Efficiency} In terms of single instruction efficiency, we found that for \texttt{2d-shape-random} layouts, \emph{designer} agents were able to modify more than three objects per turn only 14\% of the times, while this number was 20\% for \texttt{2d-shape-pattern} layouts. Thus, it further confirms that people are effective at using the patterns to optimize language for the task. We also notice how the COCO layouts elicit semantic relations that are not present in the 2D-shape layouts, so while we expect that some of the language from 2D-shape layouts will translate to real-world scenes, such as references to locations, and shapes, the realm of semantic relations might need a separate treatment.

% Figures \ref{fig:sample-dial-2D} and \ref{fig:sample-dial-COCO} show sample crowdsourced dialogs for both 2D-shape and COCO layouts respectively. Notice how the COCO layouts elicit semantic relations that are not present in the 2D-shape layouts, so while we expect that some of the language from 2D-shape layouts will translate to real-world scenes, such as references to locations, and shapes, the realm of semantic relations might need a separate treatment.

%\noindent Figure \ref{fig:sample-dial-2D} and \ref{fig:sample-dial-COCO} are sample dialogs constructed by multiple crowdsourced agents for 2D-shape and COCO layout respectively.

\section{Related Work}
\label{sec:relatedwork}

Given some of the limitations of tasks such as human-robot interactions, text-to-scene conversion or visual question answering, there has been recent interest on building more complex multimodal datasets of visually grounded dialogs~\cite{geman2015visual,mostafazadeh2017image,das2017visual,kim2017codraw}. 
% Unlike these previous works, our dataset does not directly use pixel images as the second modality but uses object layouts. 
Our goal oriented task of re-constructing the spatial distribution of objects in a canvas through conversational interactions sets our task apart from these previous works. 
%More closely related is the work of~\citet{kim2017codraw} where a drawing task is proposed between two agents in a cartoon scene world. 
In our work, we additionally explore re-construction of layouts corresponding to real-world images with a focus on the inclusion of spatial and geometric reasoning and more dynamic dialog activities while completing the tasks. 

Another aspect that sets our work apart from previous efforts in this domain is that we leverage asynchronous dialog interactions. However, there have been important previous works studying this type of dialogs in the more general setting (e.g.~\citet{blaylock2002synchronization,joty2011unsupervised,tavafi2013dialogue}). We similarly show that our proposed visually grounded task is feasible under asynchronous dialogs. 

Finally, object layouts, and visual grounding on geometric primitives has generally been of interest to study the compositionality of language. The work of~\citet{mitchell2013generating, fitzgerald2013learning} used synthetic object layouts and simple scenes to study referring expressions, while~\citet{obj2textEMNLP2017} used layouts from real images for image captioning. \citet{andreas2016neural}, and ~\citet{johnson2017clevr} introduced synthetic abstract scene datasets to test visual question answering. Our work is instead focused on visually grounded dialogs for spatial reasoning.
% \citet{suhr2017corpus} proposed a new dataset of abstract object layouts for a task of determining the veracity of a given natural language statement. Similarly ~\citet{andreas2016neural}, and ~\citet{johnson2017clevr} introduced synthetic abstract scene datasets to test visual question answering. 
% Our work is instead focused on visually grounded dialogs.

\section{Conclusions}
\label{sec:conclusions}
\vspace{-0.04in}
We developed Chat-crowd, a framework and associated platform to collect dialogs for goal-oriented tasks involving visual reasoning. Our platform incorporates mechanisms to encourage diverse dialog activities and provides a new way of evaluating the performance of crowdsourcing agents during the task. Our system demonstrated the feasibility of a \emph{director}-\emph{designer} agent interaction to reconstruct input visual layouts based only on asynchronous dialog interactions. 
% This opens the possibility to collect other more diverse visual dialog datasets through asynchronous interactions without the need to have two agents interacting at the same time. 
% \vspace{-0.15in}

\paragraph{Acknowledgment} This project was funded in part by an IBM Faculty Award to V.O.

\bibliography{naaclhlt2019}
\bibliographystyle{acl_natbib}

\end{document}